\documentclass[twoside]{article}

\usepackage[accepted]{aistats2021}
\usepackage[utf8]{inputenc} 
\usepackage[T1]{fontenc}    

\usepackage[round]{natbib}

\usepackage{hyperref}       
\usepackage{url}            
\usepackage{booktabs}       
\usepackage{amsfonts}       
\usepackage{nicefrac}       
\usepackage{microtype}      
\usepackage{balance}

\usepackage{amsmath,amsfonts,bm}

\def\eqref#1{equation~\ref{#1}}

\def\1{\bm{1}}

\def\vu{{\bm{u}}}

\def\vx{{\bm{x}}}

\def\vz{{\bm{z}}}

\def\mQ{{\bm{Q}}}

\DeclareMathAlphabet{\mathsfit}{\encodingdefault}{\sfdefault}{m}{sl}
\SetMathAlphabet{\mathsfit}{bold}{\encodingdefault}{\sfdefault}{bx}{n}

\newcommand{\E}{\mathbb{E}}

\newcommand{\R}{\mathbb{R}}

\newcommand{\KL}{\mathrm{KL}}

\DeclareMathOperator*{\argmax}{arg\,max}

\usepackage{hyperref}
\usepackage{url}
\usepackage{graphicx}

\usepackage{caption}
\usepackage{subcaption}
\usepackage[]{algorithm2e}
\usepackage{stackengine}
\usepackage{floatrow}

\usepackage{tikz}
\usetikzlibrary{bayesnet}
\usetikzlibrary{positioning}

\newcommand{\Dalpha}{{D}^\alpha}

\SetKwInput{KwInit}{Init}

\begin{document}

\twocolumn[

\aistatstitle{Explore the Context: Optimal Data Collection for Context-Conditional Dynamics Models}

\aistatsauthor{ Jan Achterhold \And Joerg Stueckler }

\aistatsaddress{ Embodied Vision Group \\
Max Planck Institute for Intelligent Systems, Tübingen, Germany\\
\texttt{\{jan.achterhold, joerg.stueckler\}@tuebingen.mpg.de}  } ]

\begin{abstract}
In this paper, we learn dynamics models for parametrized families of dynamical
systems with varying properties.
The dynamics models are formulated as stochastic processes conditioned on a latent context variable which is inferred from observed transitions of the respective system.
The probabilistic formulation allows us to compute an action sequence which, for a limited number of environment interactions, optimally explores the given system within the parametrized family.
This is achieved by steering the system through transitions being most informative for the context variable.

We demonstrate the effectiveness of our method for exploration on a non-linear toy-problem and two well-known reinforcement learning environments.
\end{abstract}

\section{INTRODUCTION}
Learning dynamics models for model-based control and model-based reinforcement learning has recently attracted significant attention by the machine learning community \citep{chua2018deep, hafner2018learning, hafner2020dream}.
In this paper, we depart from the setting of learning a single forward dynamics model per system.
Instead, we learn the dynamics of a distribution of systems which are parameterized by a latent context variable.
The latent variable is only observable through the modulations it causes in the dynamics of the systems.
Our modeling approach is based on the Neural Processes (NP) \citep{garnelo2018neural} framework: The dynamics model, which is shared across all systems we train on, can be \textit{conditioned} on \textit{context observations} from a particular system instance, similar to Gaussian Process dynamics models \citep{deisenroth2011pilco}.
Conditioning on context observations implicitly requires performing inference on the latent variables that govern the dynamics.
This process can also be interpreted as a form of meta-learning, as we learn to adapt a shared model for a specific task.
Our shared dynamics model then becomes system-specific.

Our main contribution is the proposal of an \textit{informed calibration scheme} in that framework.
We ask the question: Given a system from a family of systems (e.g., the family of pendulums with varying pole masses), what action sequence should one apply to the (unknown) system to identify it as quickly as possible within that family (e.g., to infer the mass of the pendulum's pole).
An optimization procedure seeks for transitions of the system being most informative for the latent variable, while respecting the dynamical constraints of the system.

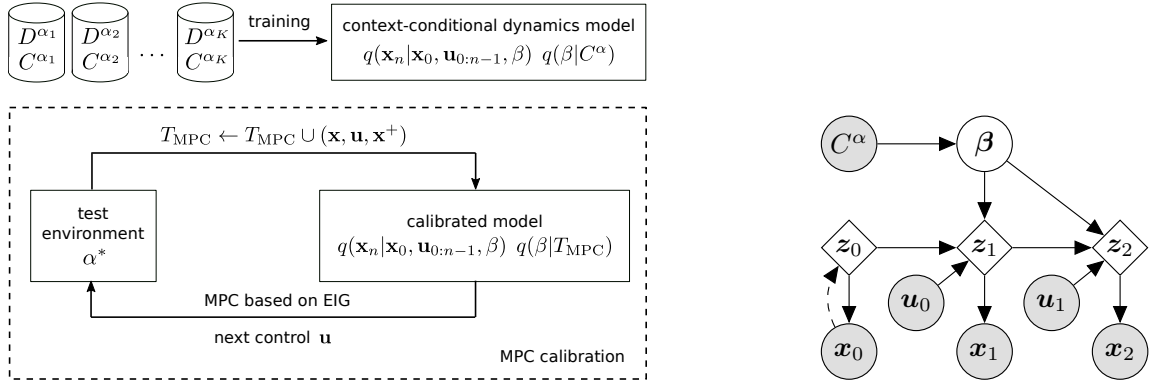
\begin{figure*}[t!]
  \centering
   \begin{subfigure}[t]{0.5\textwidth}
  \centering
  \includegraphics[width=0.99\textwidth]{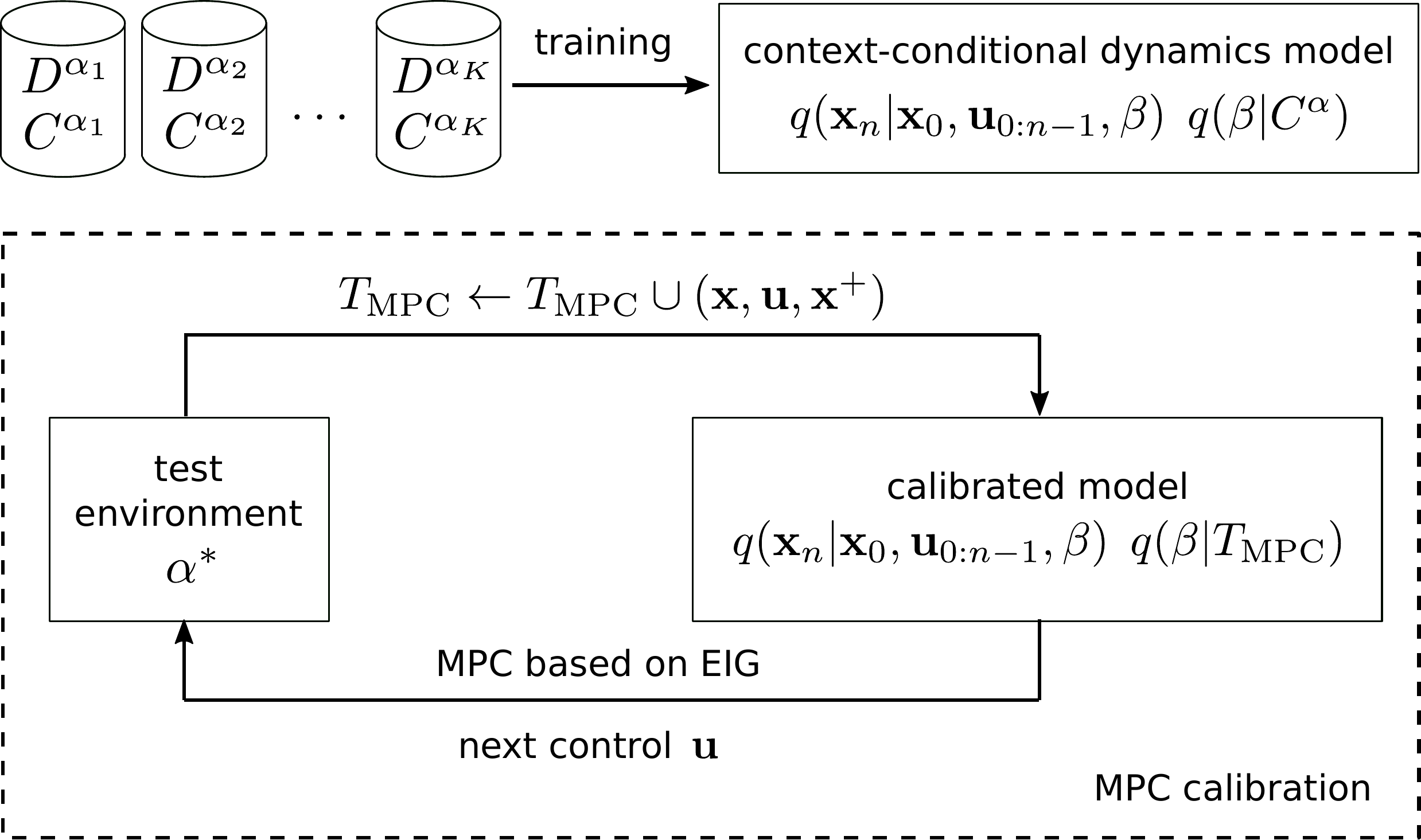}
  \caption{Overview of our proposed model training and model-predictive calibration approach.
  First, we learn a context-conditional dynamics model on rollouts from a set of $K$ parameterized environments which yield target data ($D^\alpha_1...D^\alpha_K$) and context data ($C^\alpha_1...C^\alpha_K$).
  To optimally calibrate a previously unseen test environment $\alpha^*$, we employ the context-conditional model for active data collection in a model-predictive control scheme.
  We then obtain the calibrated model by conditioning on the observations from the calibration rollout $T_{\text{MPC}}$.}
  \label{fig:method_overview}
  \end{subfigure} \hspace{1em}
  \begin{subfigure}[t]{0.45\textwidth}
  \centering
  \resizebox {!} {10em} {
    \begin{tikzpicture}[]
      \begin{scope}
        \node[det] (h0) {$\vz_0$};
        \node[det, right=of h0] (h1) {$\vz_1$};
        \node[det, right=of h1] (h2) {$\vz_2$};
        \node[latent, above=0.6 of h1](beta){$\boldsymbol{\beta}$};
        \node[obs, left=of beta] (ctx) {$C^\alpha$};
        \node[obs, below=0.6 of h0] (x0) {$\vx_0$};
        \node[obs, below=0.6 of h1] (x1) {$\vx_1$};
        \node[obs, below=0.6 of h2] (x2) {$\vx_2$};
        \node[obs, left=0.15 of h1, yshift=-0.7cm] (u0) {$\vu_0$};
        \node[obs, left=0.15 of h2, yshift=-0.7cm] (u1) {$\vu_1$};
        \edge {h0}{h1};
        \edge {h1}{h2};
        \edge {h0}{x0};
        \edge {h1}{x1};
        \edge {h2}{x2};
        \edge{beta}{h1};
        \edge{beta}{h2};
        \edge{ctx}{beta};
        \edge {u0}{h1};
        \edge {u1}{h2};
        \edge [dashed, bend left]{x0}{h0};
      \end{scope}
    \end{tikzpicture} %
  }
  \caption{Graphical model depicting the proposed context-conditional forward dynamics.
  The action-conditioned Markovian dynamics are modeled deterministically in a latent space.
  Observations $\vx_n$ of the system are random variables conditioned on their respective latent state $\vz_n$.
  The dynamics are conditioned on actions $\mathbf{u}_n$ and a latent context belief $\bm{\beta}$ which encodes an observed context set $C^\alpha$.
  For prediction, the initial latent state is inferred from the first observation $\vx_0$ (dashed line).}
  \label{fig:method_dgm}
  \end{subfigure}
  \caption{(a) overview of our proposed calibration approach and (b) context-conditional dynamics model.}
\end{figure*}

To this end, we build upon the NP formulation from \cite{garnelo2018neural} to construct a context-conditional dynamics model with a probabilistic context encoder which regresses the belief over a latent context variable (Section~\ref{sec:method}).
This allows us to formulate a calibration procedure which optimizes for a sequence of actions that minimizes uncertainty in the latent context variable.
We formulate an open-loop and model-predictive calibration algorithm to plan for the optimal sequence of actions to execute.
In Figure~\ref{fig:method_overview} we sketch an overview of our proposed calibration scheme.

On an illustrative toy environment (Section~\ref{sec:exp_toy}), we show that the uncertainty in the latent context variable corresponds to the informativeness of the set of context observations about the latent factors.

We also apply our method to a modified "Pendulum" environment from OpenAI Gym \citep{brockman2016openaigym} and a modified "MountainCar" environment \citep{moore1990efficient} which we both extended by varying properties of the underlying dynamics (Sections~\ref{sec:exp_pendulum} and \ref{sec:mountaincar}).
Our algorithm exhibits a reasonable and explainable behavior for the given task.
The calibration procedure we propose yields a calibration sequence which outperforms a random calibration sequence in terms of prediction accuracy of the conditioned dynamics model.

In summary, our contributions are as follows:
\begin{itemize}\setlength\itemsep{0ex}
    \item We apply the framework of Neural Processes \citep{garnelo2018neural} with a probabilistic context encoder to formulate a latent dynamics model.  We demonstrate in experiments that the probabilistic model yields meaningful posterior uncertainties in the context variable given observations of dynamical systems.
    \item Based on this probabilistic formulation, we develop an information-theoretic calibration scheme based on expected information gain (EIG) and model-predictive control (MPC).
    We further demonstrate that our calibration scheme outperforms a baseline which generates actions randomly.
\end{itemize}

Code and further resources to reproduce the experiments are available at \\ \url{https://explorethecontext.is.tue.mpg.de}.

\section{RELATED WORK}

\paragraph{Neural Processes} The idea of context-conditional modeling of distributions over functions using a permutation invariant context embedding was first introduced by \cite{garnelo2018conditional} as Conditional Neural Processes (CNPs).
The authors apply CNPs on image completion and classification tasks.
While \cite{garnelo2018conditional} mainly use a deterministic belief for the context encoding, also a latent variable model is introduced, but a deterministic influence of the latent context encoding on function predictions remains.
\cite{garnelo2018neural} present a more formal treatment of the latent variable model formulation coined Neural Processes (NPs).
Sequential Neural Processes \citep{singh2019sequential} extend the original Neural Processes formulation by a temporal dynamics model on the latent context embedding.
Another application of the Neural Process framework has recently been proposed for the domain of learning dynamics models for physics-based systems \citep{zhu2020neural}.
We ground our work on these results to learn a model which allows to globally reason about the uncertainty in the dynamics from the uncertainty in the context encoding, which is crucial for our proposed calibration approaches.

\paragraph{Dynamics Model Learning}
In the seminal PILCO approach~\citep{deisenroth2011pilco}, Gaussian Process (GP) dynamics models are learned for control tasks such as cart-poles.
\cite{fraccaro2017disentangled} propose an approach for modelling systems with partial observability through linear time-dependent models, with state inference performed by Kalman filtering in the latent space of a variational autoencoder.
To enable planning in the learned dynamics models, \cite{watter2015embed} learn locally linear models.
Probabilistic dynamics models in combination with planning based on the cross-entropy method \citep{rubinstein1999cross} have shown to outperform model-free reinforcement learning approaches in terms of sample efficiency \citep{chua2018deep} and can be trained directly on image representations \citep{hafner2018learning}.
These methods do not consider variations in the underlying system which can be explained through context variables.
We develop a probabilistic context-dependent dynamics model and an information-theoretic planning scheme for calibration based on model-predictive control.

\paragraph{Meta-Learning}
Neural and Gaussian Processes can be seen as types of meta learning algorithms which facilitate few-shot learning of the data distribution~\citep{garnelo2018neural,garnelo2018conditional}.
Meta learning of dynamics models has been explored in the domain of model-based reinforcement learning~\citep{nagabandi2019learning,smundsson2018_metagp}.
\cite{nagabandi2019learning} propose to combine gradient-based~\citep{finn2017model} and recurrence-based~\citep{duan2016_rl2} meta learning for online adaptation of the dynamics model.
Different to our approach, this meta learning scheme does not explicitly model the dynamics model dependent on a context variable.
Calibration on the target system requires fine-tuning the deep neural network.
We propose a deep probabilistic model and a learning scheme which allows for inferring such a context variable through calibration.
Similar to our approach \cite{smundsson2018_metagp} include a latent context variable into a probabilistic hierarchical dynamics model which they choose to model using Gaussian Processes.
The method uses probabilistic inference to determine a Gaussian context variable from data.
We use a deep encoder to regress probabilistic beliefs on the context variable and propose an information-theoretic approach for dynamics model calibration.

\paragraph{Active Learning and Exploration}
The method we propose for system calibration differs in substantial points from what is termed \textit{exploration} in reinforcement learning.
While in exploration the goal of the agent is to visit previously unseen regions of the state space for potentially finding behaviors yielding higher returns, we assume to stay in the domain of systems we observed during training.
However, some concepts from exploration approaches in reinforcement learning translate to our method, especially those based on uncertainty- and information-theoretic active learning principles~\citep{epshteyn2008active,golovin2010optimal,shyam2019model,sekar2020planning, tschantz2020reinforcement}.
\cite{buissonfenet2019actively} develop an active learning approach for GP dynamics models exploiting the GP predictive uncertainty.
Popular examples of active learning strategies are expected error reduction (EER) \citep{roy2001_eer} or expected information gain (EIG) \citep{mackay1992information}.
Our information-theoretic calibration approach is formulated based on a variant of EIG.

\paragraph{Optimal Experimental Design}
Broadly speaking, the idea of \textit{Optimal Experimental Design} is to select experiments which reveal maximal information about quantities of interest~\citep{franceschini2008model}.
An information theoretic approach based on EIG as utility function dates even back to the 1950s \citep{lindley1956measure}.
\cite{foster2019variational} propose variational approximations for EIG estimation to infer an informative sequence of experiments. Differently, we learn a context-dependent dynamics model which facilitates EIG estimation from the latent context posterior under dynamics constraints. However, their approach relates to our method as they proposed to use function approximators for amortized inference, which is similar to our context encoder, giving an amortized posterior for the latent context variable.

\section{METHOD}

\label{sec:method}
We assume that the observed data of a dynamical system is generated by the following parametrized Markovian discrete-time state-space model
\begin{align}
\label{eq:statespace}
\vx_{n+1} = f(\vx_n, \vu_n, \bm{\alpha}) + \bm{\epsilon}
\end{align}
with $n \in \mathbb{N}_0$, $\vx_n \in \mathcal{X} \subseteq \mathbb{R}^{\mathrm{dim}(\mathcal{X})}$, $\vx_0 \in \mathcal{X}_0 \subseteq \mathcal{X}$, $\vu_n \in \mathcal{U} \subseteq \mathbb{R}^{\mathrm{dim}(\mathcal{U})}$, $\bm{\alpha} \in \mathcal{A} \subseteq \mathbb{R}^{\mathrm{dim}(\mathcal{A})}$, $\bm{\epsilon} \sim \mathcal{N}(0, \mQ)$.
In the following, the term \textit{rollout of length $N$} refers to a sequence of states and actions $[\vx_0, \vu_0, \vx_1, \vu_1, ..., \vu_{N-1}, \vx_N]$, with equation~\ref{eq:statespace} holding for each \textit{transition} $(\vx_n, \vu_n, \vx_{n+1})$.

The target data $\Dalpha$ consists of rollouts of the system for fixed parameters $\bm{\alpha}$ and randomly sampled initial states $\vx_0$ and action sequences $\vu_{0:N-1} := [\vu_0, ..., \vu_{N-1}]$.
Context data $C^\alpha = \left\{ ( \vx, \vu, \vx^+ ) \right\}, \vx^+ = f(\vx, \vu, \bm{\alpha}) + \bm{\epsilon}$ is a set of transitions generated by the system with parameters $\bm{\alpha}$.
We aim at determining a probabilistic context-conditional dynamics model $p( \vx_{n} | \vx_{0:n-1}, \vu_{0:n-1}, C^\alpha)$ which maximizes the expected data log-likelihood \begin{align}
    \max \: \E_{\alpha\sim{A}} \left[ \log p(D^\alpha | C^ \alpha) \right]
\end{align}
where ${A}$ denotes the empirical distribution of parameter values $\bm{\alpha}$ and $p(D^\alpha | C^ \alpha) = \prod_{n=1}^{N} p( \vx_{n} | \vx_{0:n-1}, \vu_{0:n-1}, C^\alpha)$.
For a graphical depiction of our modeling approach, see Figure~\ref{fig:method_dgm}.

\paragraph{Latent context variable} Since we cannot directly observe $\bm{\alpha}$ and also do not know its representation, we introduce a latent variable $\bm{\beta} \in \mathbb{R}^B$ whose representation we learn and which encodes context information corresponding to $\bm{\alpha}$ contained in the context set,
\begin{equation}
    \label{eq:integral}
    \log p(D^\alpha | C^\alpha) = \log \int p(D^\alpha | \bm{\beta}) p(\bm{\beta} | C^\alpha)\,\mathrm{d}\bm{\beta}.
\end{equation}
We model $p(D^\alpha | \bm{\beta})$ by $q(D^\alpha | \bm{\beta}) = \prod_{n=1}^{N} q( \vx_{n} | \vx_{0}, \vu_{0:n-1}, \bm{\beta} )$ with \textit{transition model} $q( \vx_{n} | \vx_{0}, \vu_{0:n-1}, \bm{\beta} )$ and approximate $p(\bm{\beta} | C^\alpha)$ by $q(\bm{\beta} | C^\alpha)$, the \textit{context encoder}.

\paragraph{Transition model} To model transition sequences of length $H$ starting at index $n$, we assume non-linear dynamics parametrized by $\bm{\beta}$ and disturbed by additive zero-mean Gaussian noise $\bm{\epsilon}$
\begin{equation}
    \label{eq:multistep}
    \bm{\hat{x}}_{n+H} = h(\vx_{n}, \vu_{n}, ..., \vu_{n+H-1}, \bm{\beta}) + \bm{\epsilon}.
\end{equation}

We implement the dynamics by a recurrent GRU cell \citep{cho2014learning} which operates in an embedding space.
The encoders $g_\mathrm{state}$, $g_\mathrm{action}$ and $g_{\bm{\beta}}$ lift, respectively, the state observation $\vx_n$, action $\vu_n$ and latent context variable $\bm{\beta}$ to the embedding space in which the GRU operations are performed
\begin{align}
\begin{split}
    \vz_n &= g_\mathrm{state}(\vx_n), \\
    \vz_{n+1} &= h_\mathrm{RNN}(\vz_n, [g_\mathrm{action}(\bm{u}_n), {g_{\bm{\beta}}}(\bm{\beta})])
\end{split}
\end{align}
where $[\cdot, \cdot]$ denotes concatenation.
The decoders $d_{\mathrm{state}, \mu}$ and $d_{\mathrm{state}, \sigma^2}$ map the propagated hidden state back to a distribution on the state space
\begin{align}
     \bm{\hat{x}}_{n+H} \sim \mathcal{N}(d_{\mathrm{state}, \mu}(\vz_{n+H}), \mathrm{diag}(d_{\mathrm{state}, \sigma^2}(\vz_{n+H})).
\end{align}

\paragraph{Context encoder} The context encoder $q(\bm{\beta} | C^\alpha)$ encodes a set of transitions $C^\alpha$ into a Gaussian belief over the latent context variable
\begin{equation}
q(\bm{\beta} | C^\alpha) = \mathcal{N}\left(\bm{\beta} \:|\: {\bm{\mu}_{\bm{\beta}}} (C^\alpha), {\mathrm{diag}(\bm{\sigma^2}_{\bm{\beta}}} (C^\alpha) \right).
\end{equation}
As a first step of the encoding procedure, every transition is lifted separately into an embedding space $\mathcal{Z} = \R^Z_{\geq 0}$ using the transition encoder $g_\mathrm{trans}$, which is non-negative due to its terminal ReLU activation function.
Then, the embedded transitions are aggregated by max pooling, making the aggregated embedding invariant to the ordering of transitions in the context set $C^\alpha$, which is a common set embedding technique \citep{zaheer2017deep, garnelo2018conditional}
\begin{equation}
    g_\mathrm{trans}: \mathcal{X} \times \mathcal{U} \times \mathcal{X} \rightarrow \mathcal{Z}, \: \bm{z} = g_\mathrm{trans}(\bm{x}, \bm{u}, \bm{x}^+)
\end{equation}
\begin{align}
    [{\bm{z}_{\bm{\beta}}}]_i = \max_{(\bm{x}, \bm{u}, \bm{x}^+) \in {C^\alpha}} [g_\mathrm{trans}(\bm{x}, \bm{u}, \bm{x}^+)]_i.
\end{align}
where $[\cdot]_i$ denotes the $i$-th element of a vector.
For an empty context set, we fix ${\bm{z}_{\bm{\beta}}} = 0$.
The mean and diagonal elements of the covariance matrix of the Gaussian belief over $\bm{\beta}$ are computed from the aggregated embedding using multilayer perceptrons
\begin{align}
    {\bm{\hat{\mu}}_{\bm{\beta}}} : \mathcal{Z} \rightarrow \R^B, \quad
    {\bm{\hat{\sigma}^2}_{\bm{\beta}}} : \mathcal{Z} \rightarrow \R^B_{>0}.
\end{align}
While ${\bm{\hat{\mu}}_{\bm{\beta}}}$ follows a standard architecture, we structure ${\bm{\hat{\sigma}^2}_{\bm{\beta}}}$ to resemble the behavior of Bayesian inference of a latent variable given noisy measurements \citep{murphy2012machine}.
In this setting, adding a datapoint to the set of measurements cannot increase uncertainty over the latent variable.
Due to the non-negativity of $g_\mathrm{trans}$ and the monotonicity of the max pooling operation, this can be achieved by requiring ${\bm{\hat{\sigma}^2}_{\bm{\beta}}}(\cdot)$ to be monotonically decreasing in the sense
\begin{align}
\begin{split}
    [{\bm{\hat{\sigma}^2}_{\bm{\beta}}}(\bm{z})]_i \geq [{\bm{\hat{\sigma}^2}_{\bm{\beta}}}(\bm{z} + \bm{\Delta_z})]_i \\
    \forall \bm{z} \in \R^Z_{\geq 0}, \bm{\Delta_z} \in \R^Z_{\geq 0}, i \in \{1,..., B\}.
\end{split}
\end{align}
To model this strictly positive, monotonically decreasing function, we squash the negated output of a multilayer perceptron having non-negative weights and activations, ${g_\mathrm{non-neg}}({\bm{z}_{\bm{\beta}}})$ through a Softplus activation function $\zeta(x) = \ln(1+\exp(x))$
\begin{align}
    {\bm{\hat{\sigma}^2}_{\bm{\beta}}}({\bm{z}_{\bm{\beta}}}) = \zeta \left( -g_\mathrm{non-neg}({\bm{z}_{\bm{\beta}}}) \right).
\end{align}
We fix the scale of the latent belief by forcing the latent belief for an empty context set to the unit Gaussian $\mathcal{N}(0, \bm{I})$ with a KL divergence penalty term $\mathcal{L}_\mathrm{KL} = \KL ( q(\bm{\beta} | C^\alpha = \{\}) \Vert \mathcal{N}(0, \bm{I}) )$.

\paragraph{Evidence maximization} For general non-linear dynamics models, the integral in equation~\ref{eq:integral} cannot be solved analytically.
Similar to \cite{garnelo2018neural}, we formulate a lower bound on the log evidence
\begin{align}
\label{eq:lowerbound}
\begin{split}
& \log p(D^\alpha | C^\alpha) \geq \mathcal{J} \\
& \quad = \mathbb{E}_{\bm{\beta} \sim q(\bm{\beta} | D^\alpha \cup C^\alpha)} \left[ \mathcal{J}_\mathrm{logll} (D^\alpha,\bm{\beta}) \right] \\ & \quad \quad - \lambda_\mathrm{KL} \KL ( q(\bm{\beta} | D^\alpha \cup C^\alpha) \:||\: q(\bm{\beta} | C^\alpha) )
\end{split}
\end{align}
where $\mathcal{J}_\mathrm{logll}$ denotes the log-likelihood of a target chunk $D^\alpha=[\vx_{n}, \vu_{n}, \vx_{n+1}, ..., \vu_{n+H-1}, \vx_{n+H}]$ and $\lambda_\mathrm{KL}$ is a weighting factor for the KL divergence.
To reduce gradient variance, especially in the beginning of training, we evaluate $\mathcal{J}_\mathrm{logll} = \mathcal{J}_\mathrm{ms} + \mathcal{J}_\mathrm{ss} + \mathcal{J}_\mathrm{rec}$ as a combination of multi-step and single-step prediction log-likelihood and the immediate reconstruction likelihood in each time step.

The multi-step log-likelihood is given by
$
    \mathcal{J}_\mathrm{ms} = \sum_{h=2}^H \log q(\bm{{x}}_{n+h} | \vx_{n}, \vu_{n:n+h-1}, \bm{\beta}),
$
while the term for single-step predictions is
$
    \mathcal{J}_\mathrm{ss} = \sum_{h=0}^{H-1} \log q(\bm{{x}}_{n+h+1} | \vx_{n+h}, \vu_{n+h}, \bm{\beta}).
$
The reconstruction term is
$
    \mathcal{J}_\mathrm{rec} = \sum_{h=0}^{H} \log q({\bm{{\hat{x}}}_{n+h}=\bm{{x}}_{n+h}} | \vx_{n+h}).
$

The final loss $\mathcal{L}$ we minimize states
\begin{equation}
\mathcal{L} = -\mathcal{J} + \mathcal{L}_\mathrm{KL}.
\end{equation}
Further architectural details are given in the supplementary material.

\subsection{Computing optimal action sequences for calibration}
\label{sec:calibration}
We will now discuss algorithms to utilize the learned models from above to define optimal calibration schemes.
We refer to "calibration" as the process of inferring unknown latent variables governing the dynamics of a system, while assuming that the model is a member of the family of dynamics models seen during training.

We formulate finding optimal actions to apply to a system for calibration as a Bayesian optimal experimental design problem with an information-theoretic utility function \citep{lindley1956measure, chaloner1995bayesian}. We choose an action sequence $\vu_0, ..., \vu_{N-1}$ to maximize the expected information gain
\begin{align}
\label{eq:eig}
\begin{split}
    & \mathrm{EIG}(\vu_{0:N-1} | \vx_0, {T}_0) = \\ & \mathbb{E}_{{T} \sim q({T} | \vx_0, \vu_{0:N-1}, {T}_0)} \left[
    H[q(\bm{\beta} | {T}_0)] - H[q(\bm{\beta} | {T}_0 \cup {T})] \right]
\end{split}
\end{align}
where $\vx_0$ is the current state of the system, ${T}_0$ are already observed transitions on the system to calibrate and $H[\cdot]$ represents the entropy.
The belief over the latent context variable $q(\bm{\beta} | {T})$ after observing a set of calibration transitions ${T}$ is given by the context encoder.
The distribution of imagined rollouts of the system $q({T}|\vx_0, \vu_0, ..., \vu_{N-1}, {T}_0)$ is generated from the multi-step transition model while marginalizing out the prior belief over the latent context variable
\begin{align}
\begin{split}
    & q({T} | \vx_0, \vu_{0:N-1}, {T}_0) = \\ & \quad \int q({T} | \vx_0, \vu_{0:N-1}, \bm{\beta}) q(\bm{\beta} | {T}_0) \mathrm{d} \bm{\beta}.
\end{split}
\end{align}

We get a Monte Carlo approximation to the expectation in \eqref{eq:eig} by approximate sampling from the above distribution.
To obtain a sample, we first sample from the latent context distribution $\bm{\beta}_0 \sim q(\bm{\beta} | {T}_0)$.
Next, we form the set of calibration transitions ${T} = \{(\vx_n, \vu_n, \vx_{n+1})\}_{n=\{0,...N-1\}}$, where $\vx_n$ for all $1 \leq n \leq N$ are predictions from the learned transition model $\vx_n = \mathbb{E}[q(\vx_n | \vx_0, \vu_0, ..., \vu_{n-1}, \bm{\beta}_0)]$.

The action sequence which the $\mathrm{EIG}$ maximization yields is most informative for the latent context variable (minimizes the posterior entropy) \textit{given the a-priori belief}, i.e., taking the a-priori uncertainty about the context variable into account.

\paragraph{Open-Loop calibration} In the open-loop calibration case, we initialize the set of already observed transitions as the empty set ${T}_0 = \{ \}$.
We then optimize for a sequence of actions $\vu_0^*, ..., \vu_{N-1}^*$ to fulfill
\begin{align}
\begin{split}
    & \vu_0^*, ..., \vu_{N-1}^* = \\ & \quad \argmax_{\vu_0, ..., \vu_{N-1} \in \mathcal{U}} \mathrm{EIG}(\vu_0, ..., \vu_{N-1} | \vx_0, {T}_0 = \{ \})
\end{split}
\end{align} using the cross-entropy method \citep{rubinstein1999cross}. We call $N$ calibration horizon.

\paragraph{MPC calibration} The Open-Loop calibration scheme computes a static action sequence at the beginning of the calibration procedure.
The planned action sequence is not updated during calibration.
However, knowledge about the system obtained through already executed system interactions may be valuable to re-plan the remaining calibration actions.
Therefore, we propose a calibration method which resembles a model-predictive control scheme known from feedback control theory, which we term \textit{MPC calibration}.
For MPC calibration, we compute an Open-Loop calibration sequence \textit{at every timestep}, with ${T}_n$ containing already observed transitions (${T}_0 = \{ \}$).
From this calibration sequence, the first action is applied to the system and the resulting transition is appended to ${T}_n$, giving ${T}_{n+1}$.
This is repeated for a fixed number of timesteps $N$.
Let $\vx_n$ be the current state of the system and ${T}_n$ contain already observed transitions.
At each timestep, we optimize
\begin{align}
\begin{split}
    & \vu_n^*, ..., \vu_{n+H-1}^* = \\ & \quad \argmax_{\vu_n, ..., \vu_{n+H-1} \in \mathcal{U}} \mathrm{EIG}(\vu_n, ..., \vu_{n+H-1} | \vx_n, {T}_n),
\end{split}
\end{align}
and apply $\vu_n^*$ as next action to the system.
The planning horizon $H$ is upper bounded by the maximal planning horizon $H_\mathrm{max}$ and the remaining steps to reach the calibration horizon as $H=\min(H_\mathrm{max}, N-n)$.
Due to the repeated optimization, the MPC calibration scheme is more computationally intensive than the Open-Loop scheme.

\section{EXPERIMENTS}

We evaluate our approach for learning and calibrating context-dependent dynamics models on an illustrative toy problem, a modified Pendulum environment from OpenAI Gym and a MountainCar environment with randomly sampled terrain profiles. We present further ablation studies in the supplementary material.

\subsection{General procedure}

\paragraph{Data collection}
For each experiment, we collect random data from the respective environments. For this, we first sample environment instances for each experiment, varying in their hidden parameters $\bm{\alpha}$. The respective samples of $\bm{\alpha}$ constitute the empirical distribution $A$. Then, we simulate two rollouts of length 100 on each sampled environment instance, by applying independently sampled actions starting from a random initial state. More specific details are given in the "Data collection" paragraph of each experiment.

\paragraph{Model training}
From the pre-generated data, we sample target chunks $D^\alpha$ and context sets $C^\alpha$ for empirical loss minimization
\begin{align}
    \min \: \E_{\alpha\sim{A}, D^\alpha, C^\alpha} \left[ \mathcal{L(D^\alpha, C^\alpha)} \right],
\end{align}
where $A$ denotes the empirical distribution of parameter values $\bm{\alpha}$. We refer to the supplementary material for details on the sampling procedure.
We train all models using the Adam optimizer \citep{kingma2015adam}. We evaluate the models after training for 50k steps for the toy problem and 100k steps for Pendulum and MountainCar\footnote{Correction of version published at AISTATS 2021: We evaluate after a fixed number of steps, not at minimum validation loss.}. We perform all evaluations on 3 independently trained models per environment and model configuration. We set the KL divergence scaling factor in equation~\ref{eq:lowerbound} to $\lambda_\mathrm{KL}=5$. For more details on the training procedure, we refer to the supplementary material.

\subsection{Toy system}
\label{sec:exp_toy}
First, we introduce an illustrative toy-problem showcasing the fundamental principles of our method.
It is a discrete-time dynamical system in state-space notation
\begin{align}
\vx_{n+1} &= \begin{pmatrix}
0.8 & 0.2 \\
-0.2 & 0.8
\end{pmatrix} \vx_n + \begin{pmatrix}
\alpha \\
0
\end{pmatrix} \delta(u_n) + \bm{\epsilon}
\end{align} with $\bm{\epsilon} \sim \mathcal{N}(0, \bm{I}\cdot(0.01)^2), \bm{x}_0 \sim \mathcal{N}(0, \bm{I}), \alpha \in [-1, 1], u_n \in [-2, 2]$.
The state transition matrix leads to a damped oscillation when the system is not excited.
A non-linear function $\delta$ squashes the input $u_n$ before it is applied to the system. Inference about the parameter $\alpha$ from empirical system trajectories is only possible when at least one transition $(\vx_n, u_n, \vx_{n+1})$ is observed with $\delta(u_n) \neq 0$. As squashing functions we define $\delta^{<1}(u_n) = \max(1- |u_n|, 0)$ and $\delta^{>1}(u_n) = \max(|u_n| - 1, 0)$, with the superscript denoting which $u_n$ are mapped to non-zero values, thus, being informative for the value of $\alpha$. We train separate models for each squashing function $\delta^{<1}(u_n)$, $\delta^{>1}(u_n)$.
As we assume homoscedastic noise in this experiment, we learn a constant $d_{\mathrm{state}, \sigma^2}$ independent of $\bm{z}$.

\paragraph{Data collection} To collect training and validation data, we sample 6k values of $\alpha \sim \mathrm{Uniform}[-1, 1]$ and generate a rollout pair for each $\alpha$, with the initial state sampled from $\bm{x}_0 \sim \mathcal{N}(0, \bm{I})$ and the control inputs from $u_n \in \mathrm{Uniform}[-2, 2]$.

\begin{figure}[tb]
 \centering
     \includegraphics[width=\linewidth, trim=0 0.5em 0 0.5em, clip]{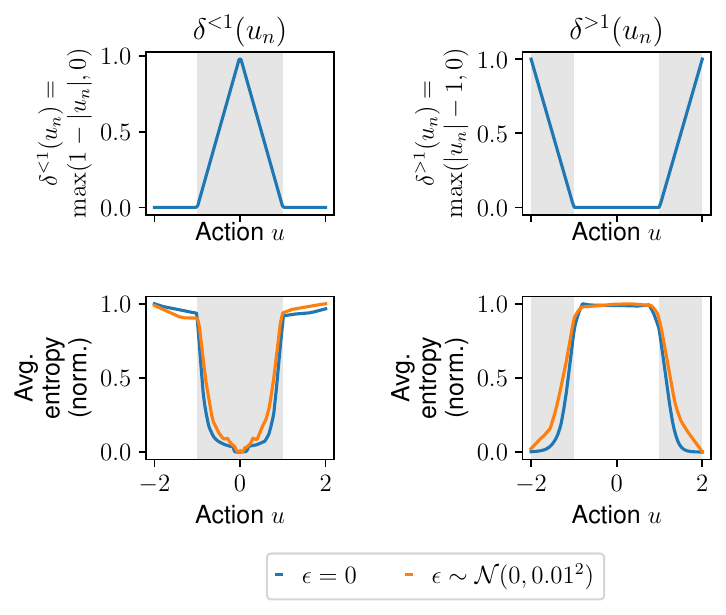}
 \caption{Characteristic behavior of the latent context belief encoder on the toy problem.
First row: Depiction of the action squashing functions $\delta^{<1}(u_n)$, $\delta^{>1}(u_n)$ (effective action magnitude).
Action regions which are informative for inferring the hidden parameter $\alpha$ are shaded in gray ($\delta(u_n) \neq 0$).
Second row: Average entropy (normalized to $[0, 1]$) of the latent context belief $H[q(\beta|C=\{\bm{x},{u},\bm{x}^+\})]$ for actions ${u}$ in systems with
Gaussian observation noise $\bm{\epsilon} \sim \mathcal{N}(0, \bm{I}\cdot(0.01)^2)$ (orange) and $\bm{\epsilon}=0$ (blue).
Non-informative actions yield a high entropy of the latent context belief, for informative actions, the entropy negatively correlates to the effective action magnitude.
Without observation noise, the entropy attains its minimum faster for increasing effective action magnitude as $\alpha$ can be inferred from low-magnitude actions with low variance. \vspace{-0.7em}}
 \label{fig:toy_entropy}
\end{figure}

\paragraph{Evaluation}
As a first evaluation, we visualize the entropy of the latent context variable for informative and non-informative context sets.
We generate random transitions $(\vx_0, u, \vx_{1})$ with $\bm{\beta} \sim q(\bm{\beta} | C^\alpha=\{\})$, $\vx_0=0$, $u_n \sim \mathrm{Uniform}[-2, 2]$ using the learned transition models for both squashing functions $\delta^{<1}(u_n)$ and $\delta^{>1}(u_n)$.
We use the learned model instead of the known ground-truth model for generating transitions to simulate the first step of a calibration procedure, in which the true model is unknown.
From each transition, we construct a single-element context set and compute the latent context belief using the learned context encoder.
In Figure~\ref{fig:toy_entropy}, we plot the entropy of the latent context belief for each action $u_n$, averaged over samples from $\bm{\beta}$.
For the posed systems with $\delta^{<1}(u_n)$ ($\delta^{>1}(u_n)$), a single-element context is informative for $\alpha$ if and only if it contains a transition with an action $|u_n|<1$ ($|u_n| > 1$).
We observe that the entropy is minimized correspondingly for our learned model which demonstrates that the context encoder predicts uncertainty well.

\begin{figure}
 \centering
     \includegraphics[width=0.99\linewidth, trim=0 0.5em 0 0.5em, clip]{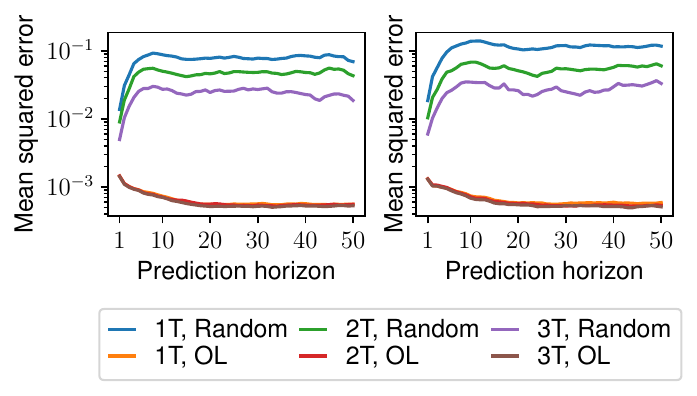}
 \caption{Evaluation of model prediction error for the toy problem. Depicted is the prediction error (lower is better) of models with random and optimal (open-loop) calibration with $\{1,2,3\}$ calibration transitions, for both action squashing schemes $\delta^{<1}(u_n)$ (left) and $\delta^{>1}(u_n)$ (right). \vspace{-1em}}
 \label{fig:toy_accuracy}
\end{figure}

In a second experiment, we evaluate the model accuracy in terms of prediction error after calibration on a set of random rollouts for 3 independently trained models.
We sample 50 system instances with $\alpha \sim \mathrm{Uniform}[-1, 1]$, and for each instance we generate 20 random rollouts.
For calibration, we limit the maximum number of transitions to $\{1,2,3\}$ and perform open-loop calibration (see section~\ref{sec:calibration}) on each system instance. As baseline, we use randomly sampled transitions for calibration.
Each calibration rollout is initialized at $\vx_0=0$, for the random calibration rollouts we uniformly sample actions from $\mathrm{Uniform}[-2, 2]$.
In Figure~\ref{fig:toy_accuracy} we depict the mean squared prediction error of the learned models for open-loop optimal and random calibration sequences of different lengths.
The prediction error is significantly lower for optimally calibrated models compared to randomly calibrated models.
As a single informative transition is sufficient to calibrate the system, MPC calibration has no advantage over Open-Loop calibration for this system.

\subsection{OpenAI Gym Pendulum}
\begin{figure*}[t!]
  \centering
  \begin{subfigure}[t]{0.49\textwidth}
  \centering
  \includegraphics[width=0.85\linewidth, trim=0 0.5em 0 0.5em, clip]{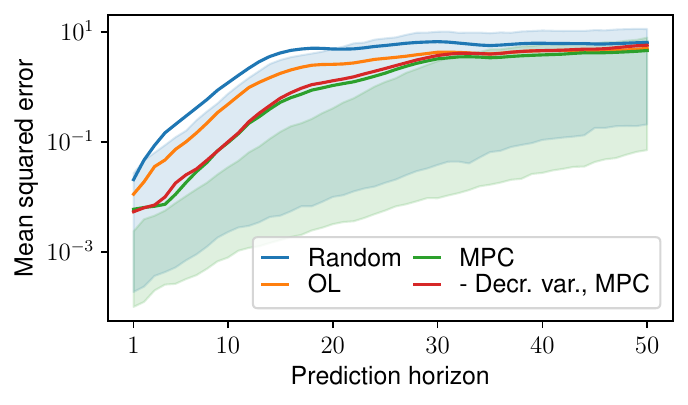}
  \caption{Pendulum}
  \label{fig:pendulum_prediction_error}
  \end{subfigure} \hfil
  \begin{subfigure}[t]{0.49\textwidth}
  \centering
  \includegraphics[width=0.85\linewidth, trim=0 0.5em 0 0.5em, clip]{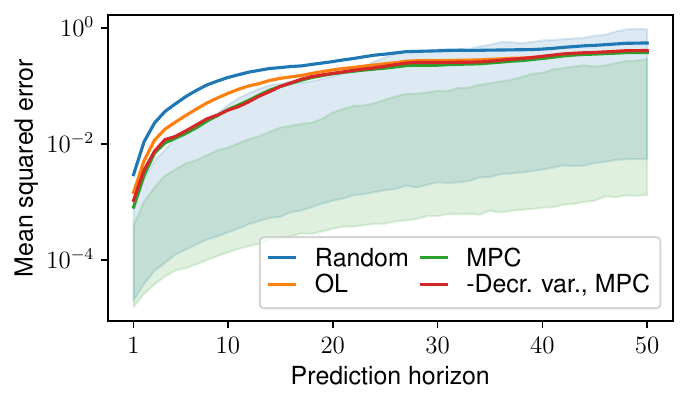}
  \caption{MountainCar}
  \label{fig:mountaincar_error}
  \end{subfigure}
  \caption{Prediction error (lower is better) of the learned (a) Pendulum and (b) MountainCar models, either conditioned on calibration data obtained with a random rollout (blue), Open-Loop calibration (orange), MPC calibration (green).
For the red curve, we train models without the strictly-decreasing variance constraint in the context encoder and perform MPC calibration.
We plot mean (line) and $20\%$ / $80\%$ quantiles (shaded area, for random and MPC calibration only for visual clarity) over 3000 rollouts.
Our proposed calibration schemes reduce prediction error compared to random calibration.
MPC calibration compares favourably to Open-Loop calibration. Enforcing the decreasing variance constraint in the context encoder slightly reduces model error after calibration for Pendulum. For MountainCar, both model variants perform similarly. Calibration rollouts contain 30 transitions for the Pendulum and 50 transitions for the MountainCar environment.\vspace{-0.5em}}
\end{figure*}

\begin{figure}[tb]
 \centering
 \begin{subfigure}[t]{0.2565\textwidth}
     \includegraphics[width=\linewidth, trim=1em 1em 0.5em 0.5em, clip]{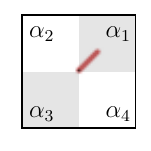}
    \caption{Pendulum dynamics vary per quadrant. The pendulum pole is depicted in red.}
    \label{fig:pendulum_quadrant}
 \end{subfigure} \hfil
 \begin{subfigure}[t]{0.67\textwidth}
     \includegraphics[width=\linewidth, trim=0.5em 1em 0 0.5em, clip]{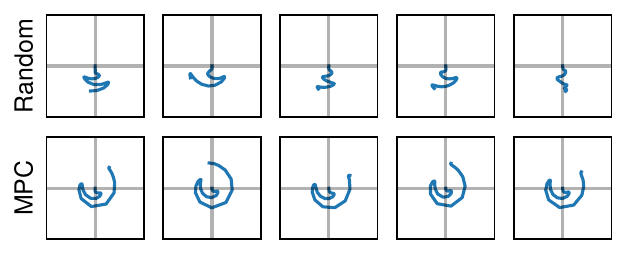}
    \caption{Pendulum angle extruded over time (starting at center) for random and MPC calibration rollouts.
 In contrast to random calibration rollouts, MPC calibration rollouts cover all quadrants to infer their dynamics.}
    \label{fig:pendulum_angle_calibration}
 \end{subfigure}
 \caption{Properties of the Pendulum environment\vspace{-0.5em}}
 \label{fig:pendulum_env}
\end{figure}

\label{sec:exp_pendulum}
In addition to the toy problem presented above, we evaluate our calibration method on a modified Pendulum environment from OpenAI Gym \citep{brockman2016openaigym}.
The Pendulum environment is a simulation of an inverted pendulum subject to gravitational force and actuable by a motor in the central rotary joint (see Figure~\ref{fig:pendulum_quadrant}). Due to torque limits, the pendulum cannot reach an upright pose without following a swing-up trajectory.

After the motor torque $u$ is clipped to a maximum magnitude of $2$, we multiply it with an angle-dependent factor $u \leftarrow u \cdot s_i \cdot \alpha_i$.
The sign $s_i$ and factor $\alpha_i$ are sampled independently for every quadrant of the pendulum system ($i \in \{1, 2, 3, 4\}$) from $\alpha_i \sim \mathrm{Uniform}[0.5, 2]$, $s_i \sim \mathrm{Uniform}\{-1, 1\}$ (see Figure~\ref{fig:pendulum_quadrant}).

\textbf{Data collection} To collect training and validation data, we first sample 110k environments with $s_i$ and $\alpha_i$ sampled as above. For each environment, we generate 2 independent rollouts with $u_n \sim \mathrm{Uniform}[-2, 2]$ and $\vx_0 \sim \mathrm{Uniform}[-\pi,\pi]\times\mathrm{Uniform}[-8,8]$, with the first dimension being the pendulum angle and the second dimension its angular velocity.

\textbf{Evaluation}
The initial state of the pendulum for calibration is sampled from $\mathrm{Uniform}[\pi-0.05,\pi+0.05]\times\mathrm{Uniform}[-0.05,0.05]$, i.e. with the pole pointing nearly downwards with small angular velocity. In contrast to the toy problem, we use the MPC calibration scheme with a CEM planning horizon of $H_\mathrm{max}=20$ and a calibration rollout length of $30$.
To approximate the expected information gain from equation~\ref{eq:eig}, we use 20 Monte Carlo samples.
We generate 50 environments for calibration with independently sampled $s_i$, $\alpha_i$. For each environment, we generate 20 random rollouts with $\vx_0$, $u_n$ sampled as in the "Data collection" paragraph.
On each environment, we apply our proposed calibration schemes and compare the predictive performance of the learned dynamics model for Open-Loop, MPC and random calibration. For random calibration, we sample random actions $u_n \sim \mathrm{Uniform}[-2, 2]$.

Figure \ref{fig:pendulum_angle_calibration} depicts the angle of the pendulum over time for random and MPC calibration rollouts.
Due to gravitational forces and inertia of the pendulum, the rollouts only cover the lower two quadrants.
Thus, with these random calibration rollouts, the dynamics in the upper two quadrants cannot be inferred.
In contrast, the MPC rollouts exhibit a swing-up behavior to reason about the dynamics in all four quadrants.
Note that this behavior solely comes from the objective to miminize entropy over the latent context variable and not through explicit modelling, e.g. via rewards.

In Figure~\ref{fig:pendulum_prediction_error} we show the prediction error of the learned dynamics model when calibrated in the Open-Loop, MPC and random scheme.
For each predicted state we compute the mean squared error to the ground-truth state (with the angle represented as $[\sin(\theta), \cos(\theta)]$) and plot its statistics (mean and quantiles) aggregated over 50 randomly sampled systems, 20 rollouts per system and 3 independently trained models, giving 3000 rollouts.
Similarly to the results from section~\ref{sec:exp_toy}, the prediction error for models calibrated with the proposed calibration schemes is lower than when calibrated with random system interventions.

\begin{figure}[tb]
 \centering
     \includegraphics[width=0.85\linewidth, trim=0 0.5em 0 0.5em, clip]{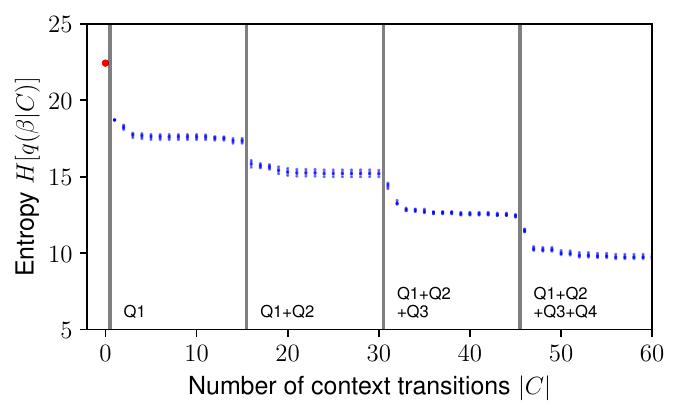}
 \caption{Entropy of the latent context belief mainly depends on the number of covered quadrants by the context set and is barely affected by the number of context transitions larger one in each quadrant.
We randomly generate 15 transitions with $u \sim \mathrm{Uniform}[-1,1]$ in all four quadrants (Q1--Q4) of several differently parameterized pendulum environments and sequentially add them to the context set, starting with quadrant 1.
Each dot corresponds to the entropy $H[q(\bm{\beta} | C)]$ of the encoded context set $C$, limited to the number of transitions given on the x-axis.
The red dot corresponds to an empty context set.
The entropy of the latent context belief saturates when transitions from already explored quadrants are added to the context set and is decreased by adding transitions from unobserved quadrants.\vspace{-0.5em}}
 \label{fig:pendulum_quadrant evaluation}
\end{figure}

As for the pendulum task in principle a single transition is sufficient to infer the dynamics in the quadrant covered by the transition, one can expect that
(1) the entropy of the latent context belief decreases for an increasing number of quadrants covered by a context set;
(2) adding transitions from an already covered quadrant to the context set does not lead to a strong reduction of entropy;
(3) the entropy is minimized when all four quadrants are covered.
Our learned context encoder features all of those properties, as we evaluate in Figure~\ref{fig:pendulum_quadrant evaluation}.
The figure depicts how the information (entropy) of the latent context belief behaves for different context sets covering different numbers of quadrants.
See Figure~\ref{fig:pendulum_quadrant evaluation}  for more details.

\paragraph{Control experiment}

In this experiment, we evaluate the performance of the calibrated models for serving as forward dynamics models in a model predictive control setting.
The task is to swing-up the Pendulum into an upright pose, which requires careful trajectory planning as, due to torque limits, the Pendulum cannot be driven to the upright pose directly.
Consequently, the calibrated dynamics model has to yield accurate predictions for successfully solving the task.
Planning is conducted using CEM with a planning horizon of 20 and a manually constructed swing-up reward function from the OpenAI Gym Pendulum implementation \citep{brockman2016openaigym}.
For all experiments, the number of transitions in a calibration rollout is set to $30$.

\begin{figure}[tb]
 \centering
     \includegraphics[width=0.85\linewidth]{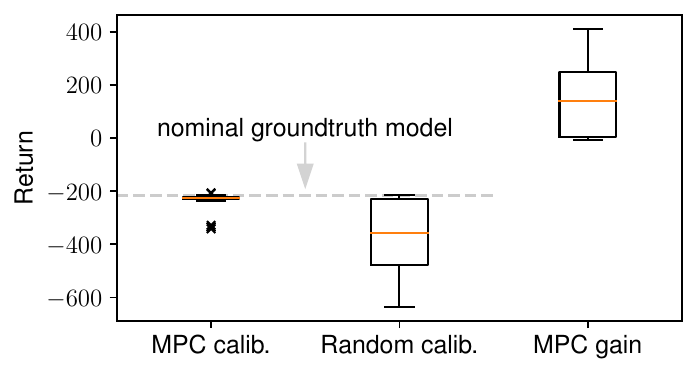}
    \caption{Swingup task cumulative reward (return) on 50 randomly sampled environment instances using a learned context-conditional model and an analytic reward function for planning.
 The learned model is conditioned on 30 transitions collected using the MPC calibration scheme or random sampling, respectively.
 "MPC gain" is gain in return when using MPC calibration scheme instead of random calibration.
 "Nominal groundtruth model" refers to the baseline performance when planning using groundtruth dynamics as forward model with $\alpha_i=1.25$.}
    \label{fig:pendulum_swingup}
\end{figure}

As depicted in Figure~\ref{fig:pendulum_swingup}, planning with a dynamics model conditioned on data collected using the MPC calibration scheme gives similar performance (return) to using a ground-truth model of the Pendulum environment.
In contrast, planning with a model conditioned on randomly sampled transitions yields a significantly lower performance.

\subsection{MountainCar environment}
\label{sec:mountaincar}
\begin{figure}[tb]
 \centering
 \begin{subfigure}[t]{0.257\textwidth}
     \includegraphics[width=\linewidth, trim=1em 1em 0.5em 0.5em, clip]{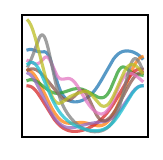}
    \caption{Sampled terrain profiles of the MountainCar environment.}
    \label{fig:mountaincar_profiles}
 \end{subfigure} \hfil
 \begin{subfigure}[t]{0.67\textwidth}
     \includegraphics[width=\linewidth, trim=0.5em 1em 0.5em 0.5em, clip]{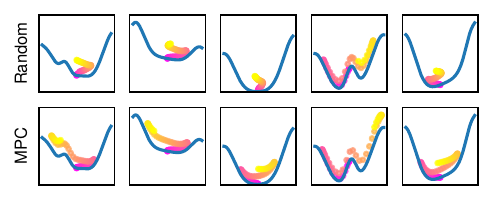}
    \caption{Random (top) vs MPC (bottom) calibration rollouts on 5 sampled MountainCar profiles (pink: t=1, yellow: t=50). The MPC rollouts cover larger extents of the profile to infer its shape. For better visibility, rollouts are extruded over time towards the positive y-direction.}
    \label{fig:mountaincar_traj}
 \end{subfigure}
 \caption{Properties of the MountainCar environment\vspace{-0.8em}}
 \label{fig:mountaincar_env}
\end{figure}

The MountainCar environment was first introduced by \cite{moore1990efficient} and is a common benchmarking problem for reinforcement learning algorithms.
In its original formulation, the task is to steer a car from a valley to a hill on a 2D profile.
Due to throttle constraints on the car, the agent has to learn to gain momentum by first steering in the opposite direction of the goal.
We extend the MountainCar environment to randomly sampled 2D terrain shapes, train context-conditional dynamics models on random rollouts and evaluate our proposed calibration routines. Figure~\ref{fig:mountaincar_profiles} illustrates randomly sampled terrain profiles. The supplementary material provides further details on the environment.

\paragraph{Data collection} For data collection, we generate random rollouts on 60k MountainCar environment instances with randomly sampled terrain profiles.

\paragraph{Evaluation} All calibration rollouts contain 50 transitions and are initialized at $x_0=\dot{x}_0=0$.
We depict the resulting random and MPC calibration rollouts in Figure~\ref{fig:mountaincar_traj}. Similar to the Pendulum environment, the MPC calibration exhibits an explorative behavior to gain information about the terrain shape. The prediction error is evaluated on $50$ randomly sampled terrain profiles, $20$ randomly generated rollouts per profile and 3 independently trained models, giving 3000 rollouts in total. The model prediction error is depicted in Figure~\ref{fig:mountaincar_error} for different calibration schemes. Again, our proposed calibration schemes compare favourably to using random transitions for calibration. The CEM planning horizon for MPC calibration for this environment is set to $H_\mathrm{max}=30$ in our experiments.

\section{CONCLUSION}

In this paper, we propose a learning method for context-dependent probabilistic dynamics models and an information-theoretic calibration approach to adapt the dynamics models to target environments from experience.
We apply the framework of Neural Processes \citep{garnelo2018neural} with a probabilistic context encoder to formulate our latent dynamics model and propose a learning approach that learns meaningful predictions of the uncertainty in the context variable.
Our calibration approach uses expected information gain in the latent context variable as optimization criterion for model-predictive control.
For evaluation, we construct an illustrative toy environment, on which we can easily distinct informative and non-informative actions for inferring a latent parameter.
Experiments on this toy problem reveal the characteristics of the probabilistic encoding for informative and non-informative actions.
We also introduce parameterized pendulum and mountain car environments to demonstrate our method on more complex systems.
On the systems, our proposed calibration methods yield models with significantly lower prediction errors compared to random calibration.
In future work, we plan to extend our method for learning dynamics models of real systems and for model-based control and reinforcement learning.
Modelling partially observable environments also is an interesting avenue of future research.

\subsubsection*{Acknowledgements}
This work has been supported by Cyber Valley and the Max Planck Society.
The authors thank the International Max Planck Research School for Intelligent Systems (IMPRS-IS)
for supporting Jan Achterhold.
We thank the anonymous reviewers for their insightful comments.

\bibliography{literature}

\end{document}


%
\runningtitle{Explore the Context: Optimal Data Collection for Context-Conditional Dynamics Models - Suppl. mat.}

%

\onecolumn
\aistatstitle{Explore the Context: Optimal Data Collection for Context-Conditional Dynamics Models - Supplementary Materials}

\aistatsauthor{ Jan Achterhold \And Joerg Stueckler }

\aistatsaddress{ Embodied Vision Group \\
Max Planck Institute for Intelligent Systems, Tübingen, Germany\\
\texttt{\{jan.achterhold, joerg.stueckler\}@tuebingen.mpg.de}  }

\section{ARCHITECTURAL DETAILS}
In the following, we describe the architectural details of our model.
Notation-wise, $[\cdot;\cdot;...]$ denotes a sequence of neural network layers.
\texttt{Linear}$(M, N)$ indicates a linear layer with $M$ input features and $N$ output features,
\texttt{ReluLinear}$(M, N)$ is a linear layer with non-negative weights $\bm{y} = \zeta(W)\bm{x}+\bm{b}$ where $\zeta(\cdot)$ is the (elementwise) ReLU function $\zeta(x) = \max(0, x)$.
\texttt{ReLU} and \texttt{Tanh} represent ReLU and hyperbolic tangent nonlinearities, respectively.
\texttt{Negate} is a negation of the input features $y=-x$.
\texttt{SoftplusOffset}($\gamma$) symbolizes a softplus nonlinearity with additive offset: $y=\ln(1+e^x)+\gamma$.

\subsection{Transition model}
The transition model consists of encoders $g_\mathrm{state}$, $g_\mathrm{action}$ and $g_{\bm{\beta}}$ to lift state (dimensionality $X$), action (dimensionality $U$) and latent context variable (dimensionality $B$) to an embedding space with dimensionality $E=200$.
A GRU cell \citep{cho2014learning} operates in the embedding space to model the dynamics.
The decoders parameterize mean and diagonal covariance on the state space given a propagated embedding.
Due to the constant additive noise assumption in the toy problem environment, in those experiments, the diagonal state space covariance of the model is learned as a constant and not modeled by a decoder.

State encoder $g_\mathrm{state}$: \\
{[\texttt{Linear}($X$, 200); \texttt{ReLU}(); \texttt{Linear}(200, $E$); \texttt{Tanh}()]}

Action encoder $g_\mathrm{action}$:\\
{[\texttt{Linear}($U$, 200); \texttt{ReLU}(); \texttt{Linear}(200, $E$); \texttt{ReLU}()]}

Latent context encoder $g_{\bm{\beta}}$:\\
{[\texttt{Linear}($B$, 200); \texttt{ReLU}(); \texttt{Linear}(200, $E$); \texttt{ReLU}()]}

GRU cell $h_\mathrm{RNN}$: \\
GRU cell with input dimension 2*$E$ (concatenation of action and latent context), state dimension $E$.

State decoder (mean) $d_{\mathrm{state}, \mu}$:\\
{[\texttt{Linear}($E$, 200); \texttt{ReLU}(); \texttt{Linear}(200, $X$)]}

State decoder (diagonal covariance) $d_{\mathrm{state}, \sigma^2}$:\\
{[\texttt{Linear}($E$, 200); \texttt{ReLU}(); \texttt{Linear}(200, $X$); \texttt{SoftplusOffset}($1\mathrm{e}^{-4}$)]}

\subsection{Context encoder}
Components of the context encoder are the transition encoder and latent context decoder networks for mean and (diagonal) standard deviation.
The dimensionality of the transition embedding space $F$ is 32 for the toy problem and 128 for all other experiments.
In ablation experiments in which we do not enforce the variance of the context encoder to be strictly decreasing with the number of context observations (referred to as "- Decr. variance"), we replace the $\texttt{ReluLinear}$ layers by standard $\texttt{Linear}$ layers.

Transition encoder:\\
{[\texttt{Linear}($2X+U$, 200); \texttt{ReLU}(); \texttt{Linear}(200, $F$); \texttt{ReLU}()]}

Latent context decoder (mean):\\
{[\texttt{Linear}($F$, 200); \texttt{ReLU}(); \texttt{Linear}(200, B)]}

Latent context decoder (diagonal standard deviation):\\
{[\texttt{ReluLinear}($F$, 200); \texttt{ReLU}(); \texttt{ReluLinear}(200, B); \texttt{Negate}(); \texttt{SoftplusOffset}($1\mathrm{e}^{-2}$)]}

\section{Training details}
Table~\ref{tab:appendix_hyperparameters} gives values for the hyperparameters used for each experiment. During training, the number of context observations for each batch element is uniformly sampled from $\{0, ..., 10\}$ for the toy problem experiment and from $\{0, ..., 50\}$ for Pendulum and MountainCar.

\begin{table}[h]
\centering
\begin{tabular}{lrrr}
\toprule
Parameter & Toy Problem & Pendulum & MountainCar \\
\midrule
Number of training steps & 50k & 100k & 100k \\
Batchsize & 64 & 512 & 512 \\
Latent context dimensionality $B$ & 1 & 16 & 16 \\
Transition embedding space dimensionality $F$ & 32 & 128 & 128 \\
\bottomrule
\end{tabular}
\caption{Hyperparameters for the toy problem, Pendulum and MountainCar experiments}
\label{tab:appendix_hyperparameters}
\end{table}

\subsection{Data sampling}
\label{sec:appendix_sampling}
As detailed in the respective environment sections, for data collection, we first randomly sample instances of the parameterized toy problem, Pendulum, and MountainCar environments.
On each environment instance we generate two independent rollouts (a \textit{rollout pair}), each with a randomly sampled initial state and randomly sampled actions.
We use rollout pairs from 5k instances of the toy problem, 100k instances of the Pendulum environment and 50k instances of the MountainCar environment as training data.
For computing a validation loss during training, we generate rollout pairs from additional 1k toy problem, 10k Pendulum and 10k MountainCar environment samples.
Enviroment instances used to evaluate the performance of the predictive model after calibration do not overlap with instances used for training and validation.

Each rollout pair we sample per environment is composed of a target rollout and a context rollout.
The target chunk $D^\alpha$ is a random, contiguous subsequence of length $50$ of the target rollout.
Each transition in the context set $C^\alpha$ is independently sampled from the target rollout with a probability $p_\mathrm{ctx-from-target}$ or from the context rollout with a probability ${p_\mathrm{ctx-from-context}=1-p_\mathrm{ctx-from-target}}$.
For the toy problem experiments, we fix $p_\mathrm{ctx-from-target}=0$.
For the Pendulum and MountainCar experiments, we set ${p_\mathrm{ctx-from-target}=0.5}$ for the first 30k training steps, then reduce it linearly to ${p_\mathrm{ctx-from-target}=0}$ until step 60k, and keep it at this value until the end of training.
We motivate this scheduling strategy to simplify the learning problem by increasing the average amount of context observations which are informative for the target chunk. When context set and target chunk are sampled from different rollouts, they may cover disjoint parts of the state space, increasing the average amount of non-informative transitions in the context set.

\subsection{Validation loss computation}
We randomly sample 5 batches from the validation data (with a batchsize of 64 for the toy problem and 512 for Pendulum and MountainCar) to construct a validation dataset (see section~\ref{sec:appendix_sampling}).
For sampling the validation batches, we fix $p_\mathrm{ctx-from-target}=0$.
The validation loss is calculated by applying the loss objective used for training (main paper, equation~14) on the validation dataset.
We report results on models yielding the lowest validation loss within the given number of training steps.

\subsection{Optimization}
We train our models using the Adam optimizer \citep{kingma2015adam} with parameters $\beta_1=0.9, \beta_2=0.999, \epsilon=1\mathrm{e}^{-4}$ and a learning rate of $1\mathrm{e}^{-3}$.
We scale gradients such that the vector of concatenated gradients has a maximal 2-norm of $1000$. As the latent context belief $p(\bm{\beta} | \cdot)$ is a multivariate Gaussian distribution with diagonal covariance matrix, the KL divergence term $\KL ( p(\bm{\beta} | D^\alpha \cup C^\alpha) \:||\: p(\bm{\beta} | C^\alpha))$ decomposes into a sum of KL divergences between scalar Gaussian distributions
\begin{equation}
\KL ( p(\bm{\beta} | D^\alpha \cup C^\alpha) \:||\: p(\bm{\beta} | C^\alpha) = \sum_i \KL ( p(\beta_i | D^\alpha \cup C^\alpha) \:||\: p(\beta_i | C^\alpha))
\end{equation}
We clip $\KL ( p(\beta_i | D^\alpha \cup C^\alpha) \:||\: p(\beta_i | C^\alpha))$ at a minimum of $0.1$ during training. This avoids local minima during the beginning of training, in which the KL divergence approaches 0 through $p(\bm{\beta} | \cdot)$ modeling a constant distribution independent of the context observations, while other loss components are not properly minimized.

\section{CEM algorithm}
\label{sec:appendix_cem}
To plan an optimal action sequence for calibration,
we use a planning algorithm based on the cross-entropy method \citep{rubinstein1999cross}
(CEM, see algorithm~\ref{algo:cem}).
We set the number of optimization iterations $T=10$, number of candidates $N_\mathrm{cand}=1000$ and number of elite candidates $N_\mathrm{elites}=100$.
The planning horizon is task dependent, during Open-Loop calibration, we use the full calibration horizon as planning horizon ($N=30$ for the Pendulum, $N=50$ for the MountainCar). During MPC calibration, the planning horizon is given by $H$ as defined in section 3.1 of the main paper.

\begin{algorithm}[h]
 \SetAlgoLined
 \KwInit{Objective function $J: \mathcal{U}^N \rightarrow \mathbb{R}, \:\: \mathcal{U} = [-u_\mathrm{max}, u_\mathrm{max}]$\\
 Maximal action magnitude $u_\mathrm{max}$, \\
 Number of optimization iterations $T$,\\ Number of candidates $N_\mathrm{cand}$, \\ Number of elites $N_\mathrm{elites}$}
 \KwResult{Optimal action sequence $(u^*_1, ..., u^*_N)$ }
 Initialize $\mu_n=0$, $\sigma_n^2=u_\mathrm{max}^2 \:\forall n \in \{1,...,N\}$\;
 \For{$t=\{1,...,T\}$}{
  \nosemic Sample $N_\mathrm{cand}$ candidate sequences \;
  \pushline\dosemic $(u^k_1, ..., u^k_N), k\in\{1,...,N_\mathrm{cand}\}$ with $\hat{u}^k_n \sim \mathcal{N}(\mu_n, \sigma^2_n)$, $u^k_n=\mathrm{clip}(\hat{u}^k_n, -u_\mathrm{max}, u_\mathrm{max})$\;
  \popline Evaluate objective $J_k = J(u^k_1, ..., u^k_N)$\;
  \nosemic Obtain set of elites $\mathcal{S}_\mathrm{elites} \subset \{1,...,N_\mathrm{cand}\} $ \;
  \pushline\dosemic with $|\mathcal{S}_\mathrm{elites}| = N_\mathrm{elites}$, $J_{k'} \geq J_k  \:\: \forall k' \in \mathcal{S}_\mathrm{elites}, k \in \{1,...,N_\mathrm{cand}\} \setminus  \mathcal{S}_\mathrm{elites}$\;
  \popline\nosemic Re-fit beliefs \;
  \pushline\dosemic $\mu_n \leftarrow \frac{1}{N_\mathrm{elites}} \sum_{k\in\mathcal{S}_\mathrm{elites}} u_n^k$\;
  $\sigma_n^2 \leftarrow \frac{1}{N_\mathrm{elites}-1} \sum_{k\in\mathcal{S}_\mathrm{elites}} (u_n^k-\mu_n)^2$\;
 }
 Set $(u^*_1, ..., u^*_N) \leftarrow (\mu_1, ..., \mu_N)$\;
 \caption{Cross-entropy method (CEM) for optimization}
 \label{algo:cem}
\end{algorithm}

\section{MountainCar environment}
The terrain profile of the MountainCar environment is generated by a linear combination of Gaussian functions $g(x; l, w) = \exp \left( -\frac{1}{2} \frac{(x-l)^2}{w^2} \right)$
\begin{align}
    y = 0.5 \cdot g(x; -1, 0.3) + 0.5 \cdot g(x; 1, 0.3) + \sum_{n=1}^N h_n \cdot g(x; l_n, w_n)
\end{align} with $x \in [-1, 1]$, $N \sim \mathrm{Uniform}\{2,...,7\}$, $h_n \sim \mathrm{Uniform}[0.1, 0.3]$, $l_n \sim \mathrm{Uniform}[-1.5, 1.5]$, $w_n \sim \mathrm{Uniform}[0.1, 0.5]$.
The Markovian state of the environment is represented by the current horizontal position and horizontal velocity.
An external tangential acceleration $u \in [-3, 3]$ can be applied to the car.
We locally approximate the dynamics of the car as a sliding block on an inclined plane with friction,
where the slope of the plane is given by the average of the profile's gradient at the current point and the gradient at the simulated next point,
akin to Heun's method for solving ordinary differential equations \citep{butcher2016numerical}.

\paragraph{Data collection}
For training and validation data collection, we generate 60k MountainCar environment instances (50k training / 10k validation) with randomly sampled terrain profiles. On each environment instance, we generate two randomly initialized ($x_0 \sim \mathrm{Uniform}[-0.8, 0.8]$, $\dot{x}_0 \sim \mathrm{Uniform}[-2, 2]$) rollouts with random actions $u_n \sim \mathrm{Uniform}[-3, 3]$.

\section{Ablation experiment: Number of calibration interactions}

\begin{figure*}[t!]
  \centering
  \begin{subfigure}[t]{0.49\textwidth}
  \centering
  \includegraphics[width=0.85\linewidth, trim=0 0.5em 0 0.5em, clip]{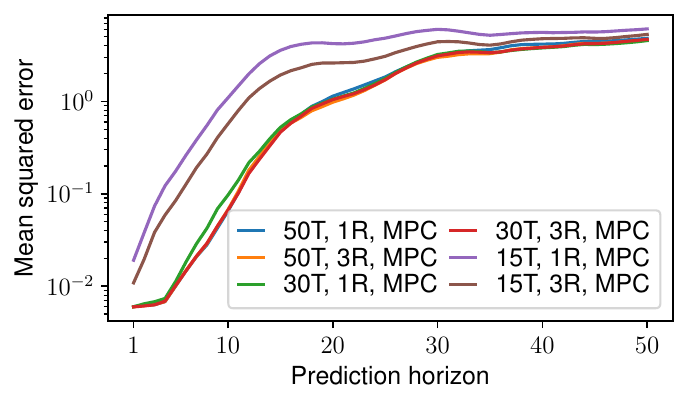}
  \caption{Pendulum}
  \end{subfigure} \hfil
  \begin{subfigure}[t]{0.49\textwidth}
  \centering
  \includegraphics[width=0.85\linewidth, trim=0 0.5em 0 0.5em, clip]{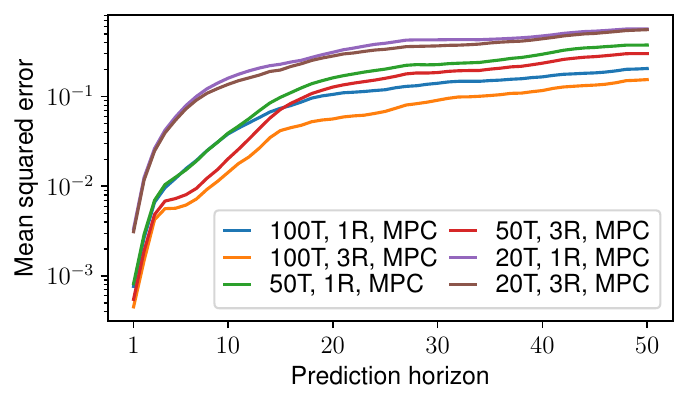}
  \caption{MountainCar}
  \end{subfigure}
  \caption{Prediction error (lower is better) of the learned (a) Pendulum and (b) MountainCar models, for MPC calibration procedures with varying number of transitions per rollout (xT) and varying number of calibration rollouts (xR). Each line represents the mean squared error over 3000 rollouts.}
  \label{fig:appendix_ablation}
\end{figure*}

As an ablation experiment, we investigate the relation between the model prediction error of a calibrated model and the number of system interactions performed during calibration.
To this end, we vary the number of calibration transitions per rollout and the number of calibration rollouts we perform for calibrating a single system.
In case of multiple rollouts, we add the transitions of previous calibration rollouts to the set of already observed system transitions $\mathcal{T}_0$ in the MPC calibration scheme.

We vary the number of transitions per rollout as $\{15, 30, 50\}$ transitions for the Pendulum environment and $\{20, 50, 100\}$ transitions for the MountainCar environment.
The number of calibration rollouts is selected from $\{1, 3\}$.
The results reported in the main paper use a single rollout with 30 transitions for the Pendulum environment and 50 transitions for the MountainCar environment.

See Figure~\ref{fig:appendix_ablation} for a depiction of the prediction error of the calibrated models.

For the Pendulum environment we observe that short calibration rollouts with 15 transitions yield significantly worse prediction results compared to rollouts of length 30 or 50, even when performing multiple calibration rollouts.
With too short rollouts, the calibration sequence can not swing-up the Pendulum to cover all (especially the upper two) quadrants.
On the other hand, longer calibration rollouts (with 50 transitions) or more calibration attempts (3 rollouts with 30 transitions) do not yield significantly better results than a single calibration rollout with 30 transitions, because all quadrants have already been covered.

For the MountainCar environment, an environment which exhibits more complex dynamics variations than the Pendulum environment, we observe that more calibration data yields models with lower prediction error for short-horizon predictions ($< 15$ steps).
However, for long-horizon predictions, a single long rollout (100 transitions) performs better than 3 short rollouts (50 transitions), although the total amount of calibration transitions is higher in the latter case.
We hypothesize that long calibration rollouts accurately explore regions which long system rollouts reach (for long prediction horizons) and thus perform better for the long-horizon case than short calibration rollouts.

\bibliography{literature}